# Deep Learning Autoencoder Approach for Handwritten Arabic Digits Recognition

Mohamed Loey, Ahmed El-Sawy
Benha University
Faculty of Computer & Informatics
Computer Science Department
Egypt
{ mloey, ahmed.el sawy }@fci.bu.edu.eg

Hazem EL-Bakry
Mansoura University
Faculty of Computer & Information Sciences
Information System Department
Egypt
helbakry5@yahoo.com

**Abstract:** *This paper presents a new unsupervised learning approach with stacked autoencoder (SAE) for Arabic handwritten digits categorization. Recently, Arabic handwritten digits recognition has been an important area due to its applications in several fields. This work is focusing on the recognition part of handwritten Arabic digits recognition that face several challenges, including the unlimited variation in human handwriting and the large public databases. Arabic digits contains ten numbers that were descended from the Indian digits system. Stacked autoencoder (SAE) tested and trained the MADBase database (Arabic handwritten digits images) that contain 10000 testing images and 60000 training images. We show that the use of SAE leads to significant improvements across different machine-learning classification algorithms. SAE is giving an average accuracy of 98.5%.*

## 1. Introduction

Pattern recognition has become a massive important due to ever demanding need of machine learning and artificial intelligence in practical problems [1]. Handwritten digits recognition is one such problem in which digits written by different authors are recognized by machines [2]. Recognition digits covers many applications such as office automation, check verification, postal address reading and printed postal codes and data entry applications are few applications [3].

Deep learning (DL) is a hierarchical structure network which through simulates the human brain's structure to extract the internal and external input data's features [4]. Deep learning based on algorithms using multilayer network such as deep neural networks, convolutional deep neural networks, deep belief networks, recurrent neural networks and stacked autoencoders. These algorithms allow computers and machines to model our world well enough to exhibit the intelligence. Autoencoders is an artificial neural network used for learning efficient encoding where the input layer have the same number of the output layer where the hidden layer has a smaller size [5,6]. In the autoencoder, the hidden layer gives a better representation of the input than the original raw input, and the hidden layer is always the compression of the input data which is the important features of the input. So, the propose of the paper is using Stacked Auto-Encoder (SAE) to create deep learning recognition system for Arabic handwritten digits recognition.

The rest of the paper is organized as follows: Section 2 gives a review on some of the related work done in the area. Section 3 describes the proposed approach, Section 4 gives an experimental result, and we list our conclusions and future work in Section 5.

## 2. Related Work

Various methods for the recognition of Latin handwritten digits [1,7,8,9] have been proposed and high recognition rates are reported. On the other hand, many researchers addressed the recognition of digits including Arabic.

In 2008, Mahmoud [11] presented a method for the automatic recognition of Arabic handwritten digits using Gabor-based features and Support Vector Machines (SVMs). He used a medium database have 21120 samples written by 44 writers. The database contain 30% for testing and the remaining 70% of the data is used for training. They achieved average recognition rates are 99.85% and 97.94% using 3 scales & 5 orientations and using 4 scales & 6 orientations, respectively.

In 2011, Melhaoui et al. [10] presented an improved technique based on Loci characteristic for recognizing Arabic digits. Their work is based on handwritten and printed numeral recognition. They trained there algorithm on dataset contain 600 Arabic digits with 400 training images and 200 testing images. They were able to achieve 99% recognition rate on this small database.

In 2013, Pandi selvi and Meyyappan [2] proposed an approach to recognize handwritten Arabic numerals using back propagation neural network. The final result shows that the proposed method provides a recognition accuracy of more than 96% for a small sample handwritten database.

In 2014, Takruri et al. [12] proposed three level classifier based on Support Vector Machine, Fuzzy C Means and Unique Pixels for the classification of handwritten Arabic digits. they tested the new algorithm on a public dataset. The dataset contain 3510 images with 40% are used for testing and 60% of images are used for training. The overall testing accuracy reported is 88%.

In 2014, Majdi Salameh [13] presented two methods about enhancing recognition rate for typewritten Arabic digits. First method that calculates number of ends of the given shape and conjunction nodes. The second method is fuzzy logic for pattern recognition that studies each shape from the shape, and then classifies it into the numbers categories. Their proposed techniques was implemented and tested on some fonts. The experimental results made high recognition rate over 95%.

In 2014, AlKhateeb et al. [14] proposed a system to classify Arabic handwritten digit recognition using Dynamic Bayesian Network. They used discrete cosine transform coefficients based features for classification. Their system trained and tested on Arabic digits database (ADBase) [3] which contains 60,000 training images and 10000 testing images. They reported average recognition accuracy of 85.26% on 10,000 testing samples.

## 3. Proposed Approach

### 3.1 Motivation

In recent years, Arabic handwritten digits recognition with different handwriting styles as well, making it important to find and work on a new and advanced solution for handwriting recognition. A deep learning systems needs a huge number of data (images) to be able to make good decisions. In [2,10-13] they applied the algorithms on a small database of handwritten images, and it's a problem in testing the variation in handwriting. In [3] proposed a large Arabic handwriting digits database called (MADBase) with 60000 training and 10000 testing images. So, MADBase database and deep learning (SAE) will be used to the suggestion of our approach.

### 3.2 Architecture

In this section, we introduce the design of SAE for digit classification. A SAE is a neural network consisting of multiple layers of sparse autoencoders in which the outputs of each layer is wired to the inputs of the successive layer [16]. A single layer in autoencoder contain three layers: (1) input layer, (2) hidden (encoding) layer, and (3) output (decoding) layer. The hidden layer try to learn to represent input layer [5,17].

### 3.2.1 Autoencoder

The main component is Autoencoder. The autoencoder is a simple three-layer neural network including an encoder and a decoder where output units are directly connected back to input units that shown in Figure 1. The proposed sparse autoencoder was trained on the raw inputs $X_n^l$, hidden layer $H_m^l$ and output layer $Y_n^l$ where n is number of inputs or outputs neuron and $m$ is number of hidden neuron and $l$ is number of sparse autoencoder. The output layer maps the input vector $I_n^l$ to the hidden layer $H_m^l$ with a non-linear function $S$:

$$H_m^l = S\left(\sum_{i=0}^{n}(W_i * X_i^l) + b_m\right) \quad (1)$$

where $W_i$ denote the parameters (or weights) associated with the connection between input unit and hidden unit. $b_m$ are a biases in hidden layer. $S(v)$ is the sigmoid function. The sigmoid function is defined as:

$$S(v) = \frac{1}{1 + e^{-v}} \quad (2)$$

The output layer $Y_n^l$ has the same number of units with the input layer and define as:

$$Y_n^l = S\left(\sum_{j=0}^{m}(\widehat{W}_j * H_j^l) + b_n\right) \quad (3)$$

where $\widehat{W}_j$ denote the parameters (or weights) associated with the connection between hidden unit and output unit. $b_n$ are the biases in output layer. $S$ is the sigmoid function shown in equation. 2.

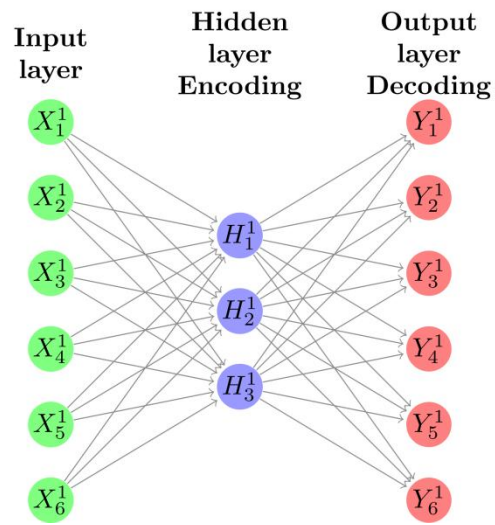

Figure 1. Sparse autoencoder structure

### 3.2.2 Stacked Autoencoder

We introduce the design of digit-level stacked autoencoder for digits classification. The first sparse autoencoder contain the input layer $X_n^l$ to learn primary features $H_m^1$ on the raw input that illustrated in Figure 2. The first sparse autoencoder produce the primary feature (feature I). The primary feature $H_m^1$ feed the input layer into the second trained sparse autoencoder that produce the secondary features (Feature II). In Figure 3 shown the primary features used as the raw input to next sparse autoencoder to learn secondary features. Then, the secondary feature treat as input layer to a softmax classifier to map secondary features to digit labels that shown in Figure 4.

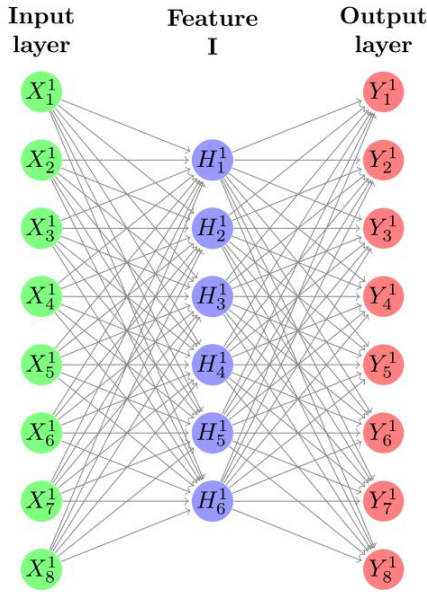

Figure 2. The first sparse autoencoder

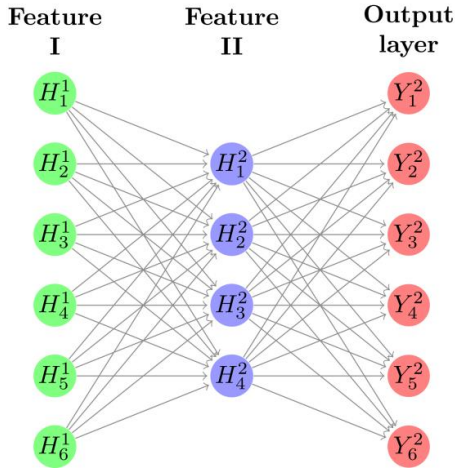

Figure 3. The second sparse autoencoder

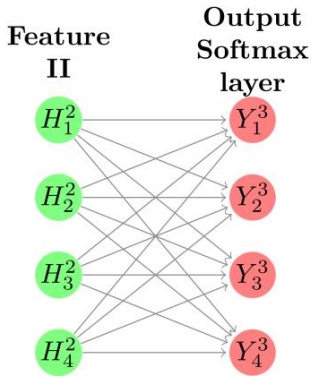

Figure 4. softmax classifier

Finally, first and second sparse autoencoder were combined with softmax classifier to produce three layers of stack autoencoder shown in Figure 5. The proposed stacked autoencoder contain two hidden layers (primary and secondary features) and the output layer (softmax classifier) capable to classify Arabic handwritten digits.

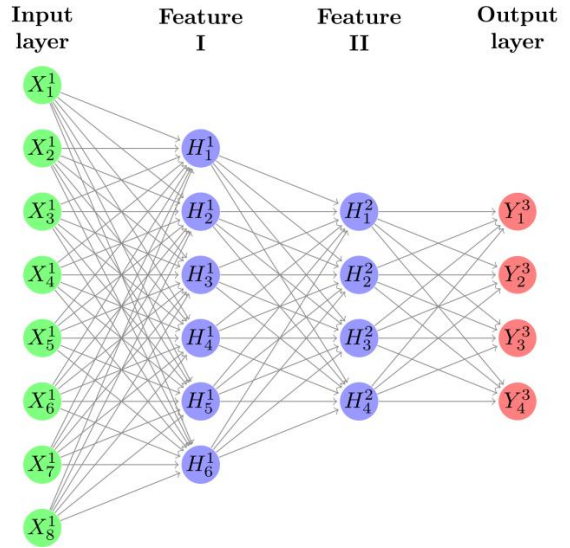

Figure 5. Proposed stacked autoencoder architecture

## 4. Experiment

### 4.1 Dataset

The MADBase database [3] based on MNIST [15] used in the experiment result. The MADBase is modified Arabic handwritten digits database contains 60,000 training images, and 10,000 test images. MADBase were written by 700 writers. Each writer wrote each digit (from 0 -9) ten times. To ensure including different writing styles, the database was gathered from different institutions: Colleges of Engineering and Law, School of Medicine, the Open University (whose students span a wide range of ages), a high school, and a governmental institution. Figure 6,7 shows samples of training and testing images of MADBase database. For researchers the MADBase is available for free and can be downloaded from (http://datacenter.aucegypt.edu/shazeem/) .

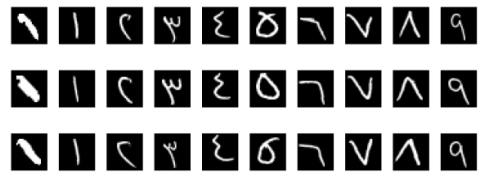

Figure 6. Training data samples from MADbase

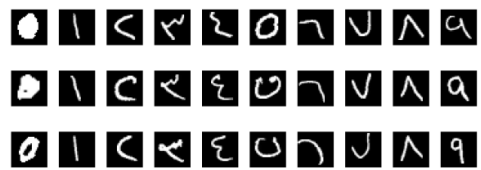

Figure 7. Testing data samples from MADbase

## 4.2 Results

In this section, we investigated the performance of the stack autoencoder. The stack autoencoder consisted of an encoder with input layers of size 784. The autoencoder discovered how to learn the features from each $28x28$ pixels of image. The experiments are conducted in MATLAB 2016a programming environment. The stacked autoencoder was done by means of the MATLAB deep learning toolbox by [18]. The first sparse autoencoder was built by input layer of size 784, hidden layer of size 392 (half of the inputs), and output layer of size 784 shown in Figure 8. The first sparse autoencoder is trained to produce 392 primary features. The second sparse autoencoder was design by input layer of size 392, hidden layer of size 196 (half of the inputs), and output layer of size 392 shown in Figure 9. The second sparse autoencoder is trained to produce 196 secondary features. The first and second sparse autoencoder uses L2 regularization to learn a sparse representation. Regularization controls the impact of an L2 regularizer for the weights of the network (and not the biases). Finally, those 196 features feed the softmax layer shown in Figure 10. The softmax layer is trained to produce ten output class.

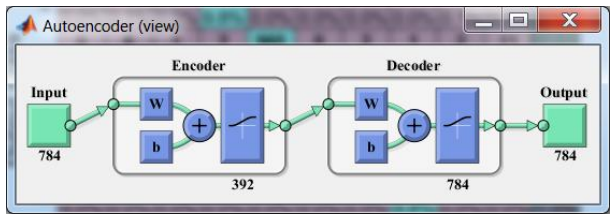

Figure 8. The proposed first sparse autoencoder

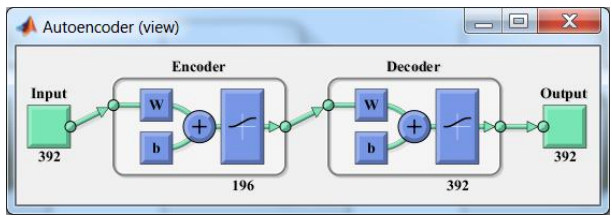

Figure 9. The proposed second sparse autoencoder

In Figure 11 shows the proposed stacked autoencoder design by one input layer of size 784, hidden layers with 392 primary feature, 196 secondary feature, and the softmax layer, and output layer with 10 labels. The proposed stacked autoencoder shown in Figure 11 train the 10000 testing data of images. To produce better results, after this phase of training is complete, fine-tuning using back-propagation can be used to improve the results by tuning the parameters of all layers are changed at the same time. we feed our stacked autoencoder with 60000 of training images. The mapping learned by the encoder part of an autoencoder can be useful for extracting features from data. Each neuron in the encoder has a vector of weights represented in Figure 12. The features learned by the autoencoder represent stroke of digits from the MADBase images.

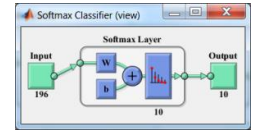

Figure 10. Softmax layer

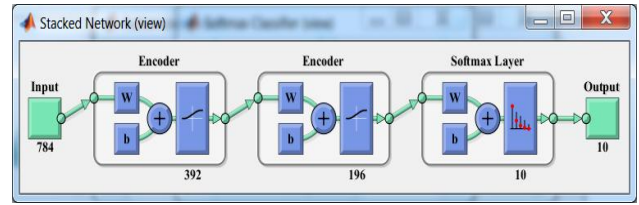

Figure 11. The proposed stacked autoencoder

The confusion matrix of the testing images shown in Figure 13, the first two diagonal cells in confusion matrix show the number and percentage of correct classifications by the trained network. For example 980 of class (1) are correctly classified as class (1). This corresponds to 9.8% of all 10000 testing images. Out of 1000 of class (1) predictions, 98.3% are correct and 1.7% are wrong. Out of 1000 of class (2) predictions, 97.1% are correct and 2.9% are wrong. Out of 1000 of class (1) cases, 98% are correctly predicted as class (1) and 2.0% are predicted as other classes. Out of 1000 of class (2), 97.9% are correctly classified as class (2) and 2.1% are classified as other classes. Overall, 98.5% of the predictions are correct and 1.5% are wrong classifications.

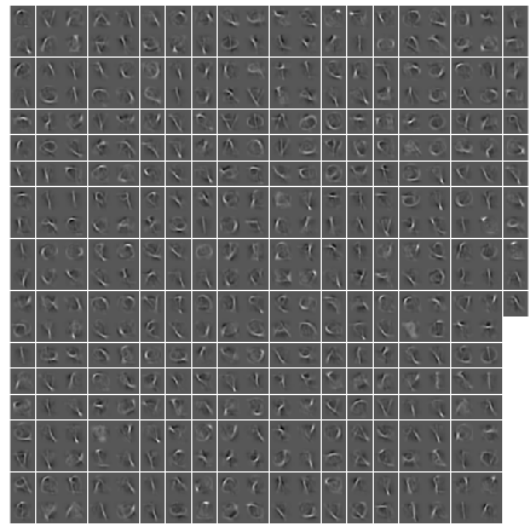

Figure 12. Visualizing the weights of the first autoencoder

Finally, in Table 1 show the obtained results with SAE on MADBase database compared with other algorithms. the proposed approach gives minimum miss-classification error compared with the results [14] in which use the same database. Although it is sometimes hard to compare the results reported in related work [2,10-13] because previous work has not experimented with large database benchmark but our results are better than most of their results. The proposed approach obtained 98.5% classification accuracy on testing data.

Figure 13. The confusion matrix of proposed stacked autoencoder

Table 1. Comparison between Proposed Approach and Other Approach

| Authors | Training Database | Testing Database | Classification Accuracy |
|---|---|---|---|
| AlKhateeb et al. [14] | 60000 | 10000 | 85.26% |
| Takruri et al. [12] | 2106 | 1404 | 88% |
| Majdi Salameh [13] | 1000 | 1000 | 95% |
| Pandi selvi and Meyyappan [2] | Samples | Samples | 96% |
| Melhaoui et al. [10] | 400 | 200 | 99% |
| Mahmoud [11] | 14784 | 6336 | 99.75% and 97.84% |
| Our Approach | 60000 | 10000 | 98.5% |

## 5. Conclusion And Future Work

In this paper, we have demonstrated the effectiveness of deep learning stacked autoencoder for Arabic handwritten digits recognition. Handwritten Recognition for Arabic numerals is an active research area which always needs an improvement in accuracy. Compare to other machine learning architectures, SAE has better performance in both images and big data of images. The purpose to use deep learning was to take advantages of the power of SAE that are able to manage large dimensions input, which allows the use of raw data inputs and able to manage large dimensions of inputs. In an experimental section we showed that the results were promising with 1.5% of testing miss classification error rate applied to the MADBase benchmark database. In the future, we will be focusing on improving the performance of handwriting Arabic digits recognition using other improved deep learning techniques and apply our approach on Arabic characters level.

## References

[1] S.S. Ali and M.U. Ghani., "Handwritten Digit Recognition Using DCT and HMMs," In Proc. Frontiers of Information Technology (FIT), 2014 12th International Conference on, pp. 303-306, 2014.

[2] P.P. Selvi and T. Meyyappan, "Recognition of Arabic numerals with grouping and ungrouping using back propagation neural network," In Proc. Pattern Recognition, Informatics and Mobile Engineering (PRIME), 2013 International Conference on, pp. 322-327, 2013.

[3] S. Abdleazeem and E. El-Sherif, "Arabic handwritten digit recognition," International Journal of Document Analysis and Recognition (IJDAR), vol. 11, no. 3, pp. 127-141, 2008.

[4] S. Ding, L. Guo, Y. Hou, "Extreme learning machine with kernel model based on deep learning," Neural Computing and Applications, pp. 1-10, 2016.

[5] J. Maria, J. Amaro, G. Falcao, L. A. Alexandre, "Stacked Autoencoders Using Low-Power Accelerated Architectures for Object Recognition in Autonomous Systems," Neural Processing Letters, vol. 43, no. 2, pp. 445-458, 2016.

[6] Q. Xu and L. Zhang, "The effect of different hidden unit number of sparse autoencoder," Proc. The 27th Chinese Control and Decision Conference (2015 CCDC), pp. 2464-2467, 2015.

[7] X.-X. Niu and C.Y. Suen, "A novel hybrid CNN–SVM classifier for recognizing handwritten digits," Pattern Recognition, vol. 45, no. 4, pp. 1318-1325, 2012.

[8] M.D. Tissera and M.D. McDonnell, "Deep extreme learning machines: supervised autoencoding architecture for classification," Neurocomputing, vol. 174, no. A, pp. 42-49, 2016.

[9] Y. Hanning and W. Peng, "Handwritten digits recognition using multiple instance learning," In Proc. Granular Computing (GrC), 2013 IEEE International Conference on, pp. 408-411, 2013.

[10] O.E. Melhaoui, M. El Hitmy, F, "Lekhal.Arabic Numerals Recognition based on an Improved Version of the Loci Characteristic," International Journal of Computer Applications, vol. 24, no.1, pp. 36-41, 2011.

[11] S.A. Mahmoud, "Arabic (Indian) handwritten digits recognition using Gabor-based features," In Proc. Innovations in Information Technology, IIT 2008. International Conference on. pp. 683-687, 2008.

[12] M. Takruri, R. Al-Hmouz, A. Al-Hmouz, "A three-level classifier: Fuzzy C Means, Support Vector Machine and unique pixels for Arabic handwritten digits," In Proc. Computer Applications & Research (WSCAR), World Symposium on. pp. 1-5, 2014.

[13] M. Salameh, "Arabic Digits Recognition Using Statistical Analysis for End/Conjunction Points and Fuzzy Logic for Pattern Recognition Techniques," World of Computer Science & Information Technology Journal, vol. 4, no. 4, pp. 50-56, 2014.

[14] J.H. AlKhateeb and M. Alseid, " DBN - Based learning for Arabic handwritten digit recognition using DCT features," In Proc. Computer Science and Information Technology (CSIT). 6th International Conference on, pp. 222-226, 2014.

[15] C.-L. Liu, K. Nakashima, H. Sako, H. Fujisawa, " Handwritten digit recognition: benchmarking of state-of-the-art techniques," Pattern Recognition, vol. 36, no. 10, pp. 2271-2285, 2003.

[16] P. Vincent, H. Larochelle, Y. Bengio, P. A. Manzagol, " Extracting and composing robust features with denoising autoencoders," Proceedings of the 25th International Conference on Machine Learning, pp. 1096-1103, 2008.

[17] P. Vincent, H. Larochelle, I. Lajoie, Y. Bengio, P. Manzagol, "Stacked Denoising Autoencoders: Learning Useful Representations in a Deep Network with a Local Denoising Criterion," J. Mach. Learn. Res, pp. 3371-3408, 2010.

[18] R. B. Palm, " Prediction as a candidate for learning deep hierarchical models of data," Technical University of Denmark. Master's thesis, 2012.